\PassOptionsToPackage{inline}{enumitem}
\documentclass[12pt]{article}
\usepackage{amsmath}
\usepackage{graphicx}
\usepackage{enumerate}
\usepackage{natbib}
\usepackage{url} 

\newcommand{\blind}{1}

\usepackage{amsfonts}
\usepackage{mathtools}
\usepackage{xcolor}
\usepackage{mathtools}
\usepackage{amssymb}
\usepackage{algorithm}
\usepackage{algorithmic}
\usepackage{enumitem}
\usepackage{booktabs}

\newtheorem{lemma}{Lemma}
\newtheorem{theorem}{Theorem}
\newtheorem{corollary}{Corollary}

\newtheorem{proposition}{Proposition}
\newtheorem{remark}{Remark}

\usepackage{tabularx}
\usepackage{graphicx}
\usepackage{caption}
\usepackage{subcaption}
\usepackage{multirow}

\newcommand{\bigo}{O}

\newcommand{\smallo}{o}
\newcommand{\sn}{S_n}
 
\newcommand{\e}{e} 
\renewcommand{\P}{\mathbb{P}} 
\DeclareMathOperator*{\E}{\mathbb{E}} 
\DeclareMathOperator*{\I}{\mathbb{I}} 
\DeclareMathOperator{\supp}{supp}
\DeclareMathOperator{\argmin}{argmin}
\DeclareMathOperator{\loglog}{loglog}
\DeclareMathOperator{\GIC}{GIC}
\DeclareMathOperator{\cov}{Cov}

\DeclarePairedDelimiter\abs{\lvert}{\rvert}%
\DeclarePairedDelimiter\norm{\lVert}{\rVert}%

\newcolumntype{Y}{>{\centering\arraybackslash}X}

\begin{document}
	
	\date{}

	\def\spacingset#1{\renewcommand{\baselinestretch}%
		{#1}\small\normalsize} \spacingset{1}

	
	\if1\blind
	{
		\title{\bf A Consistent and Scalable Algorithm for Best Subset Selection in Single Index Models}
		
		\author{Borui Tang, Jin Zhu, Junxian Zhu,
			Xueqin Wang and Heping Zhang \\
			University of Science and Technology of China, 
			Sun Yat-Sen University\\
			National University of Singapore, 
			Yale University
			\thanks{Borui Tang, Jin Zhu and Junxian Zhu contributed equally. Xueqin Wang (wangxq20@ustc.edu.cn) and Heping Zhang (heping.zhang@yale.edu) are corresponding authors.}
		}
		\maketitle
	} \fi

	\if0\blind
	{
		\bigskip
		\bigskip
		\bigskip
		\begin{center}
			{\LARGE\bf A Consistent and Scalable Algorithm for Best Subset Selection in Single Index Models}
		\end{center}
		\medskip
	} \fi
	
	\bigskip

\begin{abstract}
	Analysis of high-dimensional data has led to increased interest in both single index models (SIMs) and the best-subset selection. SIMs provide an interpretable and flexible modeling framework for high-dimensional data, while the best-subset selection aims to find a sparse model from a large set of predictors. However, the best-subset selection in
	high-dimensional models is known to be computationally intractable. { Existing proxy algorithms are appealing but do not yield the best-subset solution.  In this paper, we directly tackle the intractability by proposing a provably scalable algorithm for the best-subset selection in high-dimensional SIMs.} We directly proved the subset consistency and oracle property for our algorithmic solution, distinguishing it from other state-of-the-art support recovery methods in SIMs. The algorithm comprises a generalized information criterion to determine the support size of the regression coefficients, eliminating the model selection tuning. Moreover, our method does not assume an error distribution or a specific link function and hence is flexible to apply. Extensive simulation results demonstrate that our method is not only computationally efficient but also able to exactly recover the best-subset in various settings (e.g., linear regression, Poisson regression, heteroscedastic models). 
\end{abstract}

	\noindent%
\vfill

	\newpage
\spacingset{1.9} 

\section{Introduction}

Single index models (SIMs) are simple yet powerful models widely used in statistics and econometrics \citep{ichimura1993semiparametric,  horowitz1996direct}. In this paper, we consider a general SIM, where we do not assume that the error term is additive with the linear predictor. Formally, suppose that $ Y\in \mathbb{R} $ is a response variable and $ X=(X_1,\dots,X_p)^{\top} \in \mathbb{R}^p$ are predictors. The model is written as 
\begin{equation}\label{eqn:model_pop}
	Y=g\big({\boldsymbol{b}}^{\top}X, \e\big),
\end{equation}
where $ \boldsymbol{b} = (b_1, \dots, b_p) \in \mathbb{R}^{p} $ is the so-called index, {$g(\cdot, \cdot)$} is an unknown link function and the error term $\e\in \mathbb{R}$ is assumed to be independent of the predictors $X$. {Model~\eqref{eqn:model_pop} includes the classical SIM as a special case, which is $Y={g'(X\boldsymbol{b})} + \e$.} 
Notably, the SIM assumes that the response
depends on a linear combination of the predictors through a link function. It is a simple but flexible model, and
hence has attracted a great deal of attention  \citep{li1989regression,li1991sliced,  ichimura1993semiparametric, horowitz1996direct,  kong2007variable,  xia2009adaptive}.

Analysis of high-dimensional data makes SIMs even more appealing because it allows a
nonlinear relationship and at the same time, avoids the ``curse of dimensionality", in contrast to
linear and nonparametric regression models \citep{yang2017high, zhong2018variable}. 
{ As in linear models, the best-subset selection is also a fundamental task in SIMs. Generally, under the assumption that $\|\boldsymbol{b}\|_0 \le s$, where $\|\boldsymbol{b}\|_0 =\sum_{j=1}^p \I(b_j \neq 0)$, the best-subset selection aims to identify the index set of relevant predictors $\mathcal{A^{\star}}=\{1\le j\le p; b_j \neq 0\}$. A natural approach to this problem is to directly minimize a loss function subject to an $\ell_0$-norm constraint. However, enumerating all possible subsets is known to be NP-hard and becomes computationally infeasible in high-dimensional settings \citep{natarajan1995sparse}. 
	To address this challenge, various proxy algorithms have been proposed, including the LASSO \citep{tibshirani1996regression}, SCAD \citep{fan2001variable}, and MCP \citep{10.1214/09-AOS729}. Theoretical properties and variants of these methods have also been comprehensively studied \citep{10.1214/009053606000000281, zou2006adaptive, 10.1214/08-EJS177, zhang2008sparsity, 10.1093/biomet/asr043, 10.3150/11-BEJ410}. 
	These methods replace the $\ell_0$-norm constraint problem with either the $\ell_1$-norm or nonconvex penalties. In particular, the methods based on the $\ell_1$-norm get a sparse solution in a polynomial times. In another line of work, recent advances in statistics and machine learning have offered promising strategies to directly tackle the computational complexity of the best-subset selection. Particularly, \cite{bahmani2011greedy} design a greedy algorithm using heuristics to find an approximate solution, \cite{huang2018constructive} formulate the best-subset selection as a primal-dual active set problem and propose a constructive method, and \cite{zhu2020polynomial} develop a splicing technique with adaptive sparsity level selection. Despite their theoretical and practical importance, such direct algorithms have not yet been developed for high-dimensional SIMs.} 

{ Before presenting our method, we briefly review the related literature on proxy algorithms.
	For simultaneous estimation of both the parametric and nonparametric components {of SIM}, various methods are developed, including those by \cite{NIPS2011_30bb3825, alquier2013sparse, radchenko2015high, luo2016forward, Ganti_Rao_Balzano_Willett_Nowak_2017, 10.1214/17-EJS1329}. Due to the difficulty of estimating the link function \citep{eftekhari2021inference}, many recent approaches, including the present work, focus instead on estimating only the index vector. For example, \cite{sheng2013direction, zhang2015direction} and \cite{zhong2018variable} proposed methods based on maximizing dependence measures. However, these methods do not guarantee variable selection consistency.}

\cite{fan2022understanding} consider an implicit regularization in over-parameterized SIMs and establish the variable selection consistency, after observing that the direction of $ \boldsymbol{b} $ can be obtained from the covariance between $ Y $ and a transformation of $ X $ \citep{Brillinger2012}. Before \cite{fan2022understanding}, \cite{neykov2016l1, 7378952} propose $ \ell_1 $-regularized estimators under Gaussian predictors, and this problem has been further studied by \cite{yang2017high, JMLR:v20:18-705, pmlr-v97-wei19b, fan2022understanding} by utilizing Stein's lemma \citep{stein2004use}. However, these methods make assumptions on both predictors and response that limit their applications.

In contrast, \cite{wang2012non} and \cite{wang2015distribution} propose a rank-based method, which achieves variable selection consistency without any assumptions on the response. Specifically, they simplify the estimation problem to a regularized least square problem and provide support recovery properties for LASSO \citep{tibshirani1996regression} and non-concave penalty \citep{fan2001variable}, respectively. \cite{wang2015distribution} establish model selection consistency for SIMs in the ``$ p>n $" scenario. Their method is easy to implement, robust to the error distribution, and achieves certifiable model selection consistency. However, the scalability of this method is limited in high-dimensional cases, as the dimension is only allowed to grow polynomially with the sample size. Furthermore, the stringent irrepresentable condition is proved necessary for model selection consistency \citep{rejchel2020rank}.
\cite{rejchel2020rank} improve the growth rate of $ p $ with respect to $ n $ and relax the irrepresentable condition to a cone invertibility factor condition. However, several problems remain open, including the optimal choice of tuning parameters \citep{rejchel2020rank}.


{
	%
	%
	%
	%
	%

{ The regularization methods are appealing and widely used in high-dimensional SIMs. However, to the best of our knowledge, direct methods for the best-subset selection have not been developed in this setting. As pointed out by \cite{Bertsimas_2020}, the $\ell_0$-constraint methods have been considered as a benchmark in high-dimensional statistics. 
	In this work, inspired by the ABESS method of \cite{zhu2020polynomial}, we introduce a direct best-subset selection algorithm for high-dimensional SIMs with desirable statistical properties and provably polynomial time complexity. Specifically, we propose a {\underline{rank}-based \underline{a}daptive \underline{be}st-\underline{s}ubset \underline{s}election} (RankABESS)} algorithm, which iteratively exchanges relevant and irrelevant variables until convergence. We demonstrate the scalability of the algorithm by proving that it terminates in polynomial time with high probability, with respect to both sample size and dimensionality. To select the support size, we adopt a generalized information criterion (GIC), thereby avoiding the need for hyperparameter tuning. Furthermore, we establish desirable statistical guarantees for our algorithmic solution, including the subset selection consistency.
	}
	
	The rest of this paper is organized as follows. In Section~\ref{sec:pre}, we introduce the settings for SIMs. In Section~\ref{sec:method}, we propose a splicing algorithm, which we refer to as RankABESS, to estimate the index vector up to a constant. Theoretical results for both statistical and computational properties are presented in Section~\ref{sec:theory}. Section~\ref{sec:simulaiton} includes exhaustive numerical experiments to illustrate the empirical performance of RankABESS. 
	We provide some additional simulation results in Appendix~A. Proofs of theoretical results are deferred to Appendix~B.

	\section{Notation and Settings} \label{sec:pre}
	\subsection{Notations}
	Denote by $ \mathbf{1}_n $ the vector in $\mathbb{R}^n$ consisting of all $1$'s. When there is no ambiguity, the subscript $n$ is omitted. For any vector $ \boldsymbol{\beta} = (\beta_1, \dots, \beta_p)^{\top} \in \mathbb{R}^p $, 
	{ we define the $ \ell_q $-norm of $ \boldsymbol{\beta} $ as $ \|\boldsymbol{\beta}\|_q=(\sum_{j=1}^p\left|\beta_j\right|^q)^{1 / q} $ for $ q\in \left.\left[1,\infty \right.\right) $. Moreover, denote the $ \ell_0 $-norm of $ \beta $ as $\|\boldsymbol{\beta}\|_0=\sum_{j=1}^p \I(\beta_j \neq 0) $, where $ \I(\cdot) $ is the indicator function.}
	Let $ \mathcal{S} = \{1, \dots, p\} $ be the full index set, and for each set $ \mathcal{A} \subseteq \mathcal{S}$, denote $ \mathcal{A}^{c} = \mathcal{S}\setminus \mathcal{A} $ as the complement of $ \mathcal{A}.$ Define $ |\mathcal{A}| $  the carnality of $ \mathcal{A} $, and $ \boldsymbol{\beta}_{\mathcal{A}} = (\beta_j, j\in \mathcal{A}) \in \mathbb{R}^{|\mathcal{A}|} $. Define the support of a vector $ \boldsymbol{\beta} $ as $ \supp(\boldsymbol{\beta}) = \{j, \beta_j\neq 0\} $. For a matrix $ \boldsymbol{X} \in \mathbb{R}^{n\times p}$, define $ \boldsymbol{X}_{\mathcal{A}} = (\boldsymbol{X}_j, j \in \mathcal{A}) \in \mathbb{R}^{n \times|\mathcal{A}|} $, where $ \boldsymbol{X}_j $ denotes the $ j $-th column of $ \boldsymbol{X} $. For any covariance matrix $\boldsymbol{M}$ and index sets $\mathcal{A}$ and $ \mathcal{B}$, define $\boldsymbol{M}_{\mathcal{A}, \mathcal{B}}=({M_{i,j}})_{i\in \mathcal{A}, j \in \mathcal{B}}$. For any vector $ \boldsymbol{u} = (u_1,\dots, u_p) \in \mathbb{R}^p $ and a index set $ \mathcal{A} $, define $ \boldsymbol{u}^{\mathcal{A}} $ to be the vector in $ \mathbb{R}^{p} $ whose $ j $-th element is $ u_j $ if $ j\in \mathcal{A} $ and $ 0 $ otherwise. {For example, $ \widehat{\boldsymbol{\beta}}^{\mathcal{A}} $ is the vector whose $ j $-th element is $ \widehat{{\beta}}_j $ if $ j\in \mathcal{A} $ and $ 0 $ otherwise.}
	
	\subsection{Identification of Index Direction}
	
	Model \eqref{eqn:model_pop} is generally not identifiable. In this subsection, we discuss the identification of the index direction.
	For instance, any shift or scaling of the index vector $ \boldsymbol{b} $ can be absorbed into the link function $ g $. 
	To identify the index direction, we first assume that $ \boldsymbol{b} $ is sparse. Denote $ \mathcal{A}^{\star} = \supp(\boldsymbol{b}) $ and $ (\mathcal{A}^{\star})^c $ the index sets of relevant and irrelevant predictors, respectively. Let $ s^{\star} = |\mathcal{A}^{\star}| $ be the sparsity level. Without loss of generality, we assume that $ \E X_{j} =0 $ for all $ j\in \{1, \dots, p\} $. We further assume the following conditions:
	\begin{enumerate}[label=(A\arabic*)]
\item \label{cond:cov} Denote by $ \boldsymbol{\Sigma} $ the population covariance matrix of predictors, i.e. $ \boldsymbol{\Sigma}=\E[XX^{\top} ] $. Assume that the covariance matrix of relevant predictors $ \boldsymbol{\Sigma}_{\mathcal{A}^{\star}, \mathcal{A}^{\star}} $ is invertible. Also, without loss of generality, assume $ \Sigma_{jj} = 1 $ for all $ j\in \{1, \dots, p\} $.
\item \label{cond:linearity}
$ \E[ X_{\mathcal{A}^{\star}}|\boldsymbol{b}_{\mathcal{A}^{\star}}^{\top} {X}_{\mathcal{A}^{\star}}] $ is a linear function of $ \boldsymbol{b}_{\mathcal{A}^{\star}}^{\top} {X}_{\mathcal{A}^{\star}} $.
\end{enumerate}
Condition \ref{cond:cov} is commonly used for high-dimensional SIMs \citep{wang2015distribution, wang2012non}. It is weaker than the condition in \cite{rejchel2020rank} that the whole covariance matrix is invertible.
Condition \ref{cond:linearity} is a standard condition in the literature of sufficient dimension reduction \citep{li1989regression, li1991sliced}. It holds for the elliptical distribution such as Gaussian distribution and $t$-distribution. Moreover, as shown in \citet{hall1993almost}, this condition is not stringent for high-dimensional data because it holds approximately for many data when the dimension tends to infinity. 

The following proposition of \cite{wang2015distribution} assures that under the above conditions, the index $ \boldsymbol{b} $ is identifiable up to a constant. 

\begin{proposition} \label{prop:index}{\citep{wang2015distribution}}
Assume Conditions \ref{cond:cov} and \ref{cond:linearity} hold, then
$$ \boldsymbol{\Sigma}_{\mathcal{A}^{\star}, \mathcal{A}^{\star}}^{-1} \cov \left\{X_{\mathcal{A}^{\star}}, \I(Y < \widetilde{Y})\right\}=\kappa \boldsymbol{b}_{\mathcal{A}^{\star}}, $$ where $ (\widetilde{X}, \widetilde{Y}) $ is an independent copy of $ (X, Y) $ and
\begin{equation*}
	\kappa = \frac{\E[\I(\widetilde{Y}< Y)\boldsymbol{b}_{\mathcal{A}^{\star}}^\top X_{\mathcal{A}^{\star}}]}{\boldsymbol{b}_{\mathcal{A}^{\star}}^{\top} \boldsymbol{\Sigma}_{\mathcal{A}^{\star}, \mathcal{A}^{\star}} \boldsymbol{b}_{\mathcal{A}^{\star}}}.
\end{equation*}
\end{proposition}
{ Proposition \ref{prop:index} implies that we can identify the index $\boldsymbol{b}$ up to a scalar when $\kappa \neq 0$. Accordingly, we define $ \boldsymbol\beta^{\star} $ as our target parameter, where $ \boldsymbol\beta^{\star}_{\mathcal{A}^{\star}} = -\boldsymbol{\Sigma}_{\mathcal{A}^{\star}, \mathcal{A}^{\star}}^{-1} \cov \left\{X_{\mathcal{A}^{\star}}, \I(Y < \widetilde{Y})\right\} = -\kappa \boldsymbol{b}_{\mathcal{A}^{\star}} $ and $ \boldsymbol\beta^{\star}_{(\mathcal{A}^{c})^{\star}} = \boldsymbol{0}$.}
When $\kappa = 0,$ identifiability becomes an issue \citep{li1991sliced, wang2015distribution}. This can happen,
for example, when $g(u, \e) = u^2 + e.$ Nevertheless, we expect that $\kappa  \neq 0$ usually holds. \cite{rejchel2020rank} provide a sufficient condition on the link function to guarantee that $\kappa  \neq 0.$ 
From now on, for clarity, we assume that Conditions \ref{cond:cov} and \ref{cond:linearity} hold and $\kappa  \neq 0$. 

\section{Methodology} \label{sec:method}

Having dealt with the identifiability of $ \boldsymbol{b},$ we consider the following sample version of SIMs,
\begin{equation} \label{eqn:model}
y_i=g\left({\boldsymbol{b}}^{\top} \boldsymbol{x}_i, {\e_i}\right), i = 1,\dots, n.
\end{equation}
Denote $ \boldsymbol{y} = (y_1, \dots, y_n) $ the response vector, and $ \boldsymbol{X} = (\boldsymbol{x}_1, \dots, \boldsymbol{x}_n)^{\top} $ the sample design matrix.
We further define the rank vector $ \boldsymbol{r} = (r_1, \dots, r_n) $, where $ r_i = \sum_{i=1}^{n} \I(y_j\le y_i) $ is the rank of $ y_i $ in $ \boldsymbol{y}.$ 

{ If $ \mathcal{A}^{\star} $ is known {\it a priori}}, Proposition \ref{prop:index} leads to the following natural estimator of $ \boldsymbol{\beta}^{\star}_{\mathcal{A}^{\star}},$

\begin{equation}
\begin{aligned}
	\widehat{\boldsymbol{\beta}}_{\mathcal{A}^{\star}} & =-\Big(\frac{1}{n }\boldsymbol{X}_{\mathcal{A}^{\star}}\boldsymbol{X}_{\mathcal{A}^{\star}}\Big)^{-1}\left[\frac{1}{n} \sum_{j=1}^n \frac{1}{n} \sum_{i=1}^n (\boldsymbol{x}_i)_{\mathcal{A}^{\star}} \I\left(y_i < y_j\right)\right] \\
	& =\Big(\frac{1}{n}\boldsymbol{X}_{\mathcal{A}^{\star}}\boldsymbol{X}_{\mathcal{A}^{\star}}\Big)^{-1}\left[\frac{1}{n} \sum_{i=1}^n (\boldsymbol{x}_i)_{\mathcal{A}^{\star}} \left\{\frac{r_i}{n}-1 / 2\right\}\right]\\
	&= \underset{\supp(\boldsymbol\beta) \subseteq \mathcal{A}^{\star}}{\argmin} l_n (\boldsymbol{\beta}),
\end{aligned}
\end{equation}
where 
\begin{equation*}
l_n (\boldsymbol{\beta}) = \sum_{i=1}^{n} \left(\frac{r_i}{n} - 1/2 - \boldsymbol{x}_i^{\top} \boldsymbol{\beta}\right)^2.
\end{equation*}
The above estimation procedure leads to an oracle estimator $ \widehat{\boldsymbol{\beta}}^{o} $, where $ \widehat{\boldsymbol{\beta}}^{o}_{\mathcal{A}^{\star}} = \widehat{\boldsymbol{\beta}}_{\mathcal{A}^{\star}} $ and $ \widehat{\boldsymbol{\beta}}^{o}_{(\mathcal{A}^{\star})^c} = \boldsymbol{0}$. Statistical properties of $ \widehat{\boldsymbol{\beta}}^{o} $ are studied in Theorem \ref{thm:oracle}. 

In practice, the support of the relevant predictors is usually unknown. Thus, we formulate the estimation problem in the best-subset selection paradigm as follows,
\begin{equation}\label{eqn:opt}
\begin{aligned}
	\min_{\boldsymbol{\beta}} l_n(\boldsymbol{\beta}), \textup{ s.t. } \|\boldsymbol{\beta}\|_0 \le s.
\end{aligned}
\end{equation}
In general, the best-subset selection is known as an NP-hard problem. Recently, \cite{zhu2020polynomial} propose a splicing procedure to solve the $ \ell_0 $-constrained optimization problem efficiently with certifiable statistical properties. This motivates us to propose a rank-based splicing algorithm to solve problem (\ref{eqn:opt}). We also provide an information criterion to select the model size to avoid parameter tuning.

For simplicity of notation, denote the sample covariance matrix by $ \boldsymbol{C} = \boldsymbol{X}^{\top} \boldsymbol{X}/n $, and let $C_{ij}$ denote its $(i,j)$-th entry. Define $ \boldsymbol{q} =  (q_1, \dots, q_p)^{\top},$ where $q_j= \boldsymbol{X}_j^{\top} ( \boldsymbol{r}/n-\boldsymbol{1}/2)/n, $ 
which is the covariance between $\boldsymbol{X}_j$ and $ \boldsymbol{r}/n, j=1, \ldots, p.$ Similarly, define the correlation vector $ \boldsymbol{\rho} = (\rho_1,\dots, \rho_p)^{\top},$ where $ \rho_j = \frac{\boldsymbol{X}_j^{\top} ( \boldsymbol{r}/n-\boldsymbol{1}/2)}{\norm{\boldsymbol{X}_j}_2 \times \norm{\boldsymbol{r}/n-\boldsymbol{1}/2}_2}, j=1, \ldots, p.$
For any {candidate} best-subset $\mathcal{A} \subseteq \{1,\dots, p\}$ with carnality $ |\mathcal{A}| =s $, we call $\mathcal{A}$ and $\mathcal{A}^c$ the active set and inactive set, respectively. Given  {an}  active set $ \mathcal{A} $, we can estimate $\boldsymbol{\beta}$ of support $\mathcal{A}$ with
\begin{equation*}
\widehat{\boldsymbol{\beta}} = \underset{\supp(\beta) \subseteq \mathcal{A}}{\argmin} l_n(\boldsymbol{\beta}).
\end{equation*}
{ Furthermore, we define two measures to quantify the relative importance of variables:
\begin{enumerate}[leftmargin=*]
	\item Backward importance: For each variable $j\in \mathcal{A}$, define  $$\xi_j=l_n(\widehat{\boldsymbol{\beta}}^{\mathcal{A} \backslash\{j\}})-l_n(\widehat{\boldsymbol{\beta}}^{\mathcal{A}})=\frac{C_{j, j}}{2}(\widehat{\beta}_j)^2,$$ which represents the increase in the loss function when the $j$-th variable is removed from the active set $\mathcal{A}$.
	\item Forward importance: For each variable $j\in \mathcal{A}^c$, define $$\zeta_j=l_n(\widehat{\boldsymbol{\beta}}^{\mathcal{A}})-l_n(\widehat{\boldsymbol{\beta}}^{\mathcal{A}}+\widehat{\boldsymbol{t}}^{\{j\}})=\frac{(q_j - \boldsymbol{X}_j^{\top}\boldsymbol{X}\widehat{\boldsymbol{\beta}}/n)^2}{2C_{j,j}},$$
	where $ \widehat{\boldsymbol{t}} = \argmin_{\boldsymbol{t}} l_n(\widehat{\boldsymbol{\beta}}^{\mathcal{A}} + \boldsymbol{t}^{\{j\}}) $. This represents the decrease in the loss function when the $j$-th variable is added to the active set.
	\end{enumerate}}
	\noindent{Intuitively, variables with small backward importance ($\xi_j$) in the active set contribute little to the model, while variables with large forward importance ($\zeta_j$) in the inactive set could substantially improve the model. Based on these measures, the ``splicing" algorithm iteratively swaps the least important variables from $ \mathcal{A} $ with the most important variables from $ \mathcal{A}^c $. Formally, given a splicing size $ k \le s $, we define 
\begin{equation}\label{eq:splicing-set}
	\begin{split}
		\mathcal{S}_{k, 1} = \{j\in {\mathcal{A}}: \sum_{i \in {\mathcal{A}}} \I( \xi_j \geq \xi_i) \leq k\}, \
		\mathcal{S}_{k, 2} = \{j\in {\mathcal{A}^c}: \sum_{i \in {\mathcal{A}^c}} \I( \zeta_j \leq \zeta_i) \leq k\},
	\end{split}
\end{equation}
which represent the sets of variables to remove and add, respectively. By swapping $ \mathcal{S}_{k,1} $ and $ \mathcal{S}_{k,2} $, we obtain an updated active set $ \widetilde{\mathcal{A}} = (\mathcal{A} \setminus \mathcal{S}_{k,1}) \cup \mathcal{S}_{k,2} $ and its complement $ (\widetilde{\mathcal{A}})^c $. We then solve for the updated coefficients $ \widetilde{\boldsymbol{\beta}}=\argmin_{\supp(\beta) \subseteq \widetilde{\mathcal{A}}} l_n(\boldsymbol{\beta}) $. 
{If the reduction in the loss function, $l_n(\widehat{\boldsymbol{\beta}})-l_n(\widetilde{\boldsymbol{\beta}})$, exceeds a pre-specified threshold}, we update $\mathcal{A} $ to $ \widetilde{\mathcal{A}}$. This iterative procedure continues until the improvement in the loss function is less than the threshold. We formalize these steps in Algorithm~\ref{alg:bess}.}
\begin{algorithm}[!t]
\caption{Rank-based best-subset Selection (RankBESS)}
\label{alg:bess}
\begin{algorithmic}[1]
	\REQUIRE A dataset $\{(\boldsymbol{x}_i, y_i)\}^{n}_{i=1}$, 
	the maximum splicing size $k_{\max}$, and a threshold $\tau_s$.
	\STATE Initialize: 
	$ \mathcal{A}^0=\{j: \sum_{i=1}^p \I(|\rho_j|\le |\rho_i|) \leq s\} $ and $\boldsymbol{\beta}_{(\mathcal{A}^0)^c}^0=0 $, $ \boldsymbol{\beta}_{\mathcal{A}^0}^0=(C_{\mathcal{A}^0,\mathcal{A}^0})^{-1} \boldsymbol{q}_{\mathcal{A}^0}  $, $t \leftarrow -1$, $ \mathcal{A}^{-1} = \emptyset $.
	\WHILE{$\mathcal{A}^{t+1} \neq \mathcal{A}^{t}$} 
	\STATE $t \leftarrow t + 1$, $ \mathcal{A}^{t+1} \leftarrow \mathcal{A}^{t} $, and $L \leftarrow l_n(\boldsymbol \beta^t)$.
	\FOR{$k=1, \ldots, k_{\max}$}
	\STATE {Compute $\xi_j\leftarrow\frac{C_{j,j}}{2 }(\beta^t_j)^2$ for each $j \in \mathcal{A}^t$ and $ \zeta_j\leftarrow \frac{(q_j - \boldsymbol{X}_j^{\top}\boldsymbol{X}{\boldsymbol{\beta}^t}/n)^2}{2C_{j,j}}$ for $j \in (\mathcal{A}^t)^c$.}
	{\STATE Update the candidate active set via splicing: $\widetilde{\mathcal{A}} \leftarrow (\mathcal{A}^t \backslash \mathcal{S}_{k,1}) \cup \mathcal{S}_{k,2}$, where $\mathcal{S}_{k,1}, \mathcal{S}_{k,2}$ are computed by \eqref{eq:splicing-set}. 
		\STATE Obtain a candidate estimator $\widetilde{\boldsymbol{\beta}}$ in which $\widetilde{\boldsymbol{\beta}}_{\widetilde{\mathcal{A}}}= (\boldsymbol{C}_{\widetilde{\mathcal{A}}, \widetilde{\mathcal{A}}})^{-1}\boldsymbol{q}_{\widetilde{\mathcal{A}}}, \; \widetilde{\boldsymbol{\beta}}_{(\widetilde{\mathcal{A}})^c}=0.$}
	\IF {$L - l_n(\widetilde{\boldsymbol \beta})>\tau_s$}
	\STATE $L  \leftarrow l_n(\widetilde{\boldsymbol \beta} )$, $({\mathcal{A}^{t+1}} , {\boldsymbol \beta}^{t+1}) \leftarrow (\widetilde{\mathcal{A}}, \widetilde{\boldsymbol\beta})$, and break from the \textbf{for} loop.
	\ENDIF
	\ENDFOR
	\ENDWHILE
	\ENSURE $(\boldsymbol \beta^{t},   \mathcal{A}^{t})$.
\end{algorithmic}
\end{algorithm}
\begin{algorithm}[!t]
\caption{Rank-based Adaptive best-subset Selection (RankABESS)}
\label{alg:abess}
\begin{algorithmic}[1]
	\REQUIRE A dataset set $\{(\boldsymbol{x}_i, y_i)\}^{n}_{i=1}$ and the maximum support size $s_{\max}$.
	\FOR {$s = 1, \ldots, s_{\max}$}
	\STATE $(\widehat{\boldsymbol \beta}_{s}, \widehat{\mathcal{A}}_{s}) \leftarrow \textbf{RankBESS(s)}\left( \{(\boldsymbol{x}_i, y_i)\}^{n}_{i=1}, k_{\max}, \tau_s \right)$.
	\ENDFOR
	\STATE Compute the size of support set that minimizes the GIC: $\widehat{s} \leftarrow \arg\min\limits_s \textup{GIC}(\widehat{\mathcal{A}}_s)$.
	\ENSURE $(\widehat{\boldsymbol \beta}_{\widehat{s}}, \widehat{\mathcal{A}}_{\widehat{s}} )$.
\end{algorithmic}
\end{algorithm}

In practice, we determine the support size $ s $ by a data driven procedure. Specifically, we use a generalized information criterion (GIC), defined as
\begin{equation*}
{\GIC(\mathcal{A}) = n \min_{\supp(\beta) \subseteq \mathcal{A}}\log l_n(\boldsymbol{\beta}) + |\mathcal{A}| \log p \loglog n,}
\end{equation*}
to identify the true model. Intuitively, we penalize the model complexity with $ \log p $ and prevent underfitting with the slow diverging rate $ \loglog n $. Given a maximum support size $ s_{\max} $, we apply the above algorithm for $ s = 1,2,\dots, s_{\max} $ and select the support size such that the $ \GIC $ attains the minimum. We state this procedure formally in Algorithm \ref{alg:abess}. The consistency of  this selection procedure will be established in Theorem \ref{thm:gic} in the next section. A practical choice of $s_{\max}$ will be provided following Theorem \ref{thm:gic}.

\begin{remark}
The key difference between ABESS in \cite{zhu2020polynomial} and {RankABESS} is that the latter uses the ranks of the response values. This is a very useful technique in robust statistics \citep{rejchel2020rank} and stabilizes the estimation for potentially heavy-tailed responses. Like ABESS, {RankABESS} retains the scalability for SIMs.
\end{remark}

\section{Theory} \label{sec:theory}
{We establish the subset selection consistency} of RankABESS in Theorems \ref{thm:supprecovery} and \ref{thm:gic}, and present the computational property of the algorithm in Theorem \ref{thm:time}. It is noteworthy that our theory is developed for the algorithmic solution directly and requires no assumption on the error distribution. We will show that our proposed method is certifiably computationally efficient, consistent, and robust to the error distribution.

For notation clarity, let $b^{\star}= \min\limits_{j \in \mathcal{A}^*}(\beta_j^*)^2$ denote the minimum signal strength.
{ We introduce the spectrum restricted condition (SRC) for the covariance matrix. A covariance matrix $\boldsymbol{M} $ satisfies $\operatorname{SRC}(s, m_s, M_s)$  for some constants $m_s$ and $M_s$, if 
$$m_s \leq \lambda_{\min} (\boldsymbol{M}_{\mathcal{A}, \mathcal{A}}) \leq \lambda_{\max} (\boldsymbol{M}_{\mathcal{A}, \mathcal{A}})\leq M_s, \quad \forall \mathcal{A} \text{ with }  \abs{\mathcal{A}} \le s. $$	}
Below, we describe several conditions that are used in our theoretical results:

\begin{enumerate}[label=(C\arabic*)]
\item \label{cond:src} $\boldsymbol \Sigma$ satisfies $\operatorname{SRC}(2s, m^0_{2s}, M^0_{2s})$ for some constants $m^0_{2s}$ and $M^0_{2s}$.

{ The SRC is commonly assumed for the sample covariance matrix in related literature; see, for example, \cite{zhang2008sparsity, huang2018constructive} and \cite{zhu2020polynomial}. In this work, we impose the SRC on the population covariance matrix to account for the randomness of the predictors. Intuitively, the SRC provides spectral bounds for the diagonal submatrices of the covariance matrix. 
	Compared with the irrepresentable condition required for RankLASSO \citep{rejchel2020rank}, {the SRC is easier to interpret and more convenient in practice \citep{zhang2008sparsity}.} A detailed comparison between these two conditions can be found in \cite{zhang2008sparsity}, which also provides two sufficient conditions for the SRC in Propositions 1 and 2. For example, the SRC can be implied by a mutual coherence condition. The superior performance of RankABESS over RankLASSO under high-correlation settings also demonstrates that RankABESS requires less restrictive assumptions on predictors to achieve exact support recovery in practice (see Section \ref{sec:simulaiton}). 
}

Condition \ref{cond:src} also implies that the spectrum of the off-diagonal submatrices of the covariance matrix can be upper bounded. Specifically, let $\nu^0_s$ be the smallest number such that the following inequality holds:
$$\lambda_{\max}(\boldsymbol \Sigma_{\mathcal{A}, \mathcal{B}})\le v^0_s, \quad \forall \abs{\mathcal{A}} \le s, \abs{\mathcal{B}} \le s, \mathcal{A}\cap \mathcal{B} = \emptyset.$$
It follows from Lemma 20 in \cite{huang2018constructive} that 
$ \nu^0_s \le \max\{1-m^0_{2s}, M^0_{2s}-1\}.$

\item \label{cond:technical} For some small constant $\delta_s >0$, let $m_s = m_s^0-\delta_s,$ $M_s = M_s^0 + \delta_s$, and $\nu_s = \nu_s^0 + \delta_s$. Assume that there exists some $\Delta>0$ such that
\begin{equation*}
	\phi_s = \frac{2(1+\frac{1+\delta_s}{1-\delta_s})^2 M_s((1+\Delta)  \frac{\nu_s}{m_s}(1+\frac{\nu_s}{m_s}))^2}{(1-\Delta)(m_s - \frac{\nu_s^2}{m_s})} < 1.
\end{equation*}

This technical condition puts some restrictions on the spectrum bounds. This is an analog of Assumption 3 in \cite{zhu2020polynomial}. A Similar condition is also assumed in \cite{huang2018constructive}.  As discussed in Remark 2 of \cite{zhu2020polynomial}, a sufficient condition can be given by $\{m_{2s} \ge 0.8123, M_{2s} \le 1.8777 \} $. Here, we introduce the constant $ \delta_s $ to control the deviation of the sample covariance matrix from the population version. See Lemma 2 in the appendix for technical details. Besides, it controls the trade-off between the stringency of the assumption and the magnitude of tail probability in our theory.

\item \label{cond:subgaussian} Assume that $x_{1j}$ is univariate subgaussian with coefficient $\omega_j$ for all $j \in( \mathcal{A}^{\star})^c$, and $(\boldsymbol{x}_1)_{\mathcal{A^{\star}}}$ is joint subgaussian with coefficient $\omega_0$. Let $\omega = \max \{\omega_j, \omega_0, j \in ( \mathcal{A}^{\star})^c\}$.

Condition \ref{cond:subgaussian} is standard in the literature for high-dimensional random design settings  \citep{ravikumar2011high,rejchel2020rank}. Although there are conditions on the distribution of predictors, we do not impose any assumption on the error distribution. In contrast, \cite{yang2017high} and \cite{fan2022understanding} require assumptions on both predictors and response; that is, $ \E[y_1^4] < \infty$ and $ \boldsymbol{x}_{1} $ follows a known distribution with density $ p_0 $ in addition to a moment condition on the score function of $ \boldsymbol{x}_1.$

\item \label{cond:threshold} $ \tau_s = \Theta(\frac{s\log p \loglog n + \sqrt{n \log p}}{n}) $.

Condition \ref{cond:threshold} characterizes the magnitude of the threshold that can simultaneously eliminate unnecessary iterations (with the lower bound) and distinguish signals from the random error (with the upper bound). Note that the second term dominates if the sample size is relatively large and vice versa. In particular, if $\frac{s^2\log p(\log\log n)^2}{n} = o(1)$, $\tau_s = \Theta(\sqrt{\frac{\log p}{n}}).$

\item \label{cond:order} $ \frac{ s^{2}\log p}{n} = \smallo(1) $.

Condition \ref{cond:order} characterizes the maximum growth rate for the support size and the dimension. It is similar to Condition 5 in \cite{zhu2020polynomial} though stronger since a sparser subset is required here to ensure that the sample covariance matrix is close enough to the population version. Similar restrictions are required in \cite{rejchel2020rank} for RankLASSO as well. However, this condition is weaker than that in \cite{fan2013tuning} for generalized linear models.

\item \label{cond:order_signal} $ {b^{\star}} = \Omega(\frac{s\log p \loglog n + \sqrt{n \log p}}{n}) $.

Condition \ref{cond:order_signal} characterizes the strength of the minimum signals. Similar conditions are commonly assumed for consistent variable selection in high-dimensional scenarios \citep{fan2013tuning, zhu2020polynomial}.

\item \label{cond:order_gic} $ \frac{(s^\star)^2 \log p \loglog n }{n} = \smallo(1) $.

Similar to Condition \ref{cond:order}, Condition \ref{cond:order_gic} ensures that the order of sparsity level and dimension is not too large. It guarantees the subset selection consistency under the GIC. 
\item \label{cond:mutual} $ \norm{\boldsymbol{\Sigma}_{\mathcal{A}^{\star}, \{j\}}}_2 \times  \sqrt{s^{\star}} \le 1 - \delta_s $, for all $ j\notin \mathcal{A}^{\star} $.

Condition \ref{cond:mutual} imposes a restriction on the correlation between relevant predictors and irrelevant ones. This is weaker than Condition (A2*) in \cite{huang2018constructive}, where a uniform upper bound for the off-diagonal elements of the correlation matrix is assumed. In contrast, we only assume the norm of specific $ {|\mathcal{A}^{\star}}| $-length vector is bounded.

\end{enumerate}

Theorem \ref{thm:supprecovery} guarantees the consistent subset selection for a given support size by showing that the true active set can be recovered with high probability, and the recovering probability tends to $ 1 $ as the sample size increases. It also lays the foundation for the subset selection consistency under the GIC and guarantees the computational efficiency.
\begin{theorem} \label{thm:supprecovery}

Let $ \widehat{\mathcal{A}} $ be the estimated active set output by Algorithm \ref{alg:bess} for an $ s \ge s^{\star} $. Assume that Conditions (C1), (C2), and (C4)-(C6) hold with $s$, and that Condition (C3) holds. Then
$$
\P (\mathcal{A}^* \subseteq \widehat{\mathcal{A}} ) = 1- \gamma(s;n,p),
$$
where
$
\gamma(s;n,p) = \bigo\left(p\exp\left\{-n K_{s,1} \frac{b^{\star}}{s}\lor \sqrt{ \frac{b^{\star}}{s}}\right\}\right) + \bigo \left(p^2\exp\{- K_{s,2}\frac{n}{2s^2}\}\right)
$
for some constants $ K_{s,1} $ and $ K_{s,2} $. Asymptotically,
$$
\lim_{n\rightarrow \infty} \P (\mathcal{A}^* \subseteq \widehat{\mathcal{A}} ) = 1.
$$
Especially, if $ s=s^{\star} $, then we have
$$
\lim_{n\rightarrow \infty} \P (\mathcal{A}^* = \widehat{\mathcal{A}} ) = 1.
$$
\end{theorem}
\begin{remark}\label{rem:supprecovery}
{While \cite{zhu2020polynomial} demonstrate a support recovery property for ABESS in a linear model, we face extra difficulties here due to the interdependence of the ranks of the responses. Furthermore, we make no assumptions about the error distribution. Whereas the support recovery results in \cite{wang2015distribution}  presuppose that $ s^{\star} = \bigo(n^{c_1}) $ and $ p = \smallo(n^{(c_2-2c_1)k}) $ for some $ 0\le c_1<1/3$, $2c_1<c_2\le 1 $ and some positive integer $ k $, our results are more lenient in this regard (C5). For the theoretical minimizer of RankLASSO, \cite{rejchel2020rank} find a comparable rate. In contrast to their approach, we directly explore aspects of the algorithmic solution here.}

\end{remark}

Theorem \ref{thm:gic} theoretically justifies our support size selection procedure via the GIC. Specifically, our model achieves the subset selection consistency as the sample size increases.
\begin{theorem}\label{thm:gic}
Assume that Conditions (C1), (C2), and (C4)-(C6) hold with $ s_{\max} $, and that Conditions (C3) and (C7) hold. Then, for $s_{\max} > s^{\star}$, under the GIC, 
with probability $ 1-\bigo(p^{-\alpha}) $ for some positive constant $ \alpha $ and a sufficiently large $ n $, the true active set is selected, that is, $ \widehat{\mathcal{A}} = \mathcal{A}^* $.
\end{theorem}
\begin{remark}
{To achieve support recovery, it is necessary to have $s_{\max} > s^{\star}$. Given the upper bound on $s^{\star}$ assumed in Condition (C7), we recommend specifying the maximum support size as $s_{\max}= \Theta(\min\{\sqrt{\frac{n}{\log{p}\loglog{n}}}, p\})$ 
	to satisfy this requirement.
	}\end{remark}
	
	\begin{remark} 
Model selection consistency for RankLASSO is studied in \cite{wang2015distribution}. However, they only allow $p$ increases in a polynomial rate with $ n $. Moreover, there are no theoretical guarantees for the tuning parameter selection procedure. In this regard, \cite{rejchel2020rank} propose two modified versions: adaptive RankLASSO and thresholded RankLASSO. Although theoretical guidance on tuning parameter selection is provided, it still relies on unknown constants, and the optimal choice of tuning parameter remains unclear. In contrast, we provide a GIC to select the support size with certifiable statistical properties and no need for parameter tuning. 
\end{remark}
Theorem \ref{thm:gic} directly implies that RankABESS leads to the oracle estimator with high probability. We further establish the asymptotic properties of the oracle estimator in the following theorem.
\begin{theorem} \label{thm:oracle}
Assume that the conditions of Theorem \ref{thm:gic} hold. Let $ \widehat{\boldsymbol{\beta}}_{\widehat{s}} $ be the output of Algorithm \ref{alg:abess}. We have
\begin{equation*}
	P\left(\widehat{\boldsymbol{\beta}}_{\widehat{s}}=\widehat{\boldsymbol{\beta}}^o\right)=1-O\left(p^{-\alpha}\right),
\end{equation*}
for some constant $ \alpha>0 $. Moreover, the following asymptotic properties are achieved.
\begin{enumerate}[leftmargin=*]
	\item (\textbf{Consistency}) $ \norm{\widehat{\boldsymbol{\beta}}_{\mathcal{A}^{\star}} - \boldsymbol\beta^{\star}_{\mathcal{A}^{\star}} }_2 = O_p(\sqrt{\frac{s^{\star}}{n}})  $.
	\item (\textbf{Asymptotic Normality}) For any $ \boldsymbol{v} \in \mathbb{R}^p $ such that $ \norm{\boldsymbol{v}}_2\le 1 $ and $ \boldsymbol{v}^{\top} \boldsymbol\Omega \boldsymbol{v} \rightarrow \Lambda $ as $ n\rightarrow \infty $ for some $\Lambda> 0 $ , where $ \boldsymbol\Omega {=  \cov [\boldsymbol \Sigma^{-1}_{\mathcal{A}^{\star}, \mathcal{A}^{\star}}(F(Y) - 1/2 - X^{\top} \boldsymbol{\beta}^{\star})X] } $, we have
	\begin{equation*}
		\sqrt{n} \boldsymbol{v}^{\top}(\widehat{\boldsymbol{\beta}}_{\mathcal{A}^{\star}} - \boldsymbol\beta^{\star}_{\mathcal{A}^{\star}} ) \stackrel{L}{\longrightarrow} N(0, \Lambda), \quad \text { as } n \rightarrow \infty.
	\end{equation*}
\end{enumerate}
\end{theorem}
Similar theoretical results for the oracle estimator are developed in \cite{wang2012non}. In comparison, we obtain a better rate at the cost of assuming subgaussianity for relevant predictors. Specifically, the $ l_2 $ error bound in probability $ O_p(\sqrt{\frac{s^{\star}}{n}}) $ is tighter than that of \cite{wang2012non}. Furthermore, for the asymptotic normality, we require $ s^{\star}=o(n^{1/2}) $, which is weaker than that in \cite{wang2012non}, where $ s^{\star}=o(n^{1/4}) $ is assumed.  Importantly, a direct implication of Theorem \ref{thm:oracle} is that our estimator enjoys post selection consistency, as shown in the following corollary.
\begin{corollary}
If the conditions of Theorem \ref{thm:gic} hold, we have $ \norm{\widehat{\boldsymbol{\beta}}_{\widehat{s}} - \boldsymbol\beta^{\star}}_2 = o_p(1) $.
\end{corollary}

Following the notation in Algorithm \ref{alg:bess}, we now turn to the computational properties of RankABESS including an upper bound for the number of iterations of our algorithm in Lemma \ref{lem:iter} and a theoretical guarantee for the polynomial complexity in Theorem \ref{thm:time}.
\begin{lemma} \label{lem:iter}
Assume that Conditions (C1)-(C4) hold and $ s\ge s^{\star} $. Then with probability at least $ 1-\gamma(s;n,p) $, at iteration $t,$ we have $ \mathcal{A}^{\star} \subseteq \mathcal{A}^t $ if
\begin{equation*}
	t > \log_{\frac{1}{\phi_s}}\left(\frac{\sn}{n(1-\Delta)(m_s - \frac{\nu_s^2}{m_s^2})b^{\star}}\right),
\end{equation*}
where $ \sn = \norm{\boldsymbol{r}/n - \mathbf{1}/2}_2^2 = \Theta(n).$
\end{lemma}
\begin{theorem} \label{thm:time}
Assume Conditions (C1) - (C7) hold. Then,  with probability $ 1 - \bigo(p^{-\alpha})$ for some positive constant $\alpha$, the computational complexity of RankABESS for a given $ s_{\max} $ is
\begin{align*}
	{O\left(s_{\max}\log \frac{\sn}{\sqrt{n \log p}}  +\frac{n\sn}{\log p \loglog n} \right)\times(nps_{\max} + ns_{\max}^2 + k_{\max} p s_{\max}),}
\end{align*}
where $ S_n $ is defined in Lemma \ref{lem:iter}.
\end{theorem}
Theorem \ref{thm:time} implies that our algorithm achieves a polynomial complexity with high probability. Intuitively, for $ s\ge s^{\star} $, Lemma \ref{lem:iter} implies that our algorithm recovers the true support within finite iterations. For $ s<s^{\star} $, the lower bound in Condition \ref{cond:threshold} ensures that {unnecessary } iterations can be avoided and the computational complexity is also reasonably bounded.
\section{Simulation} \label{sec:simulaiton}
\begin{figure}
\centering
\includegraphics[width=0.9\linewidth]{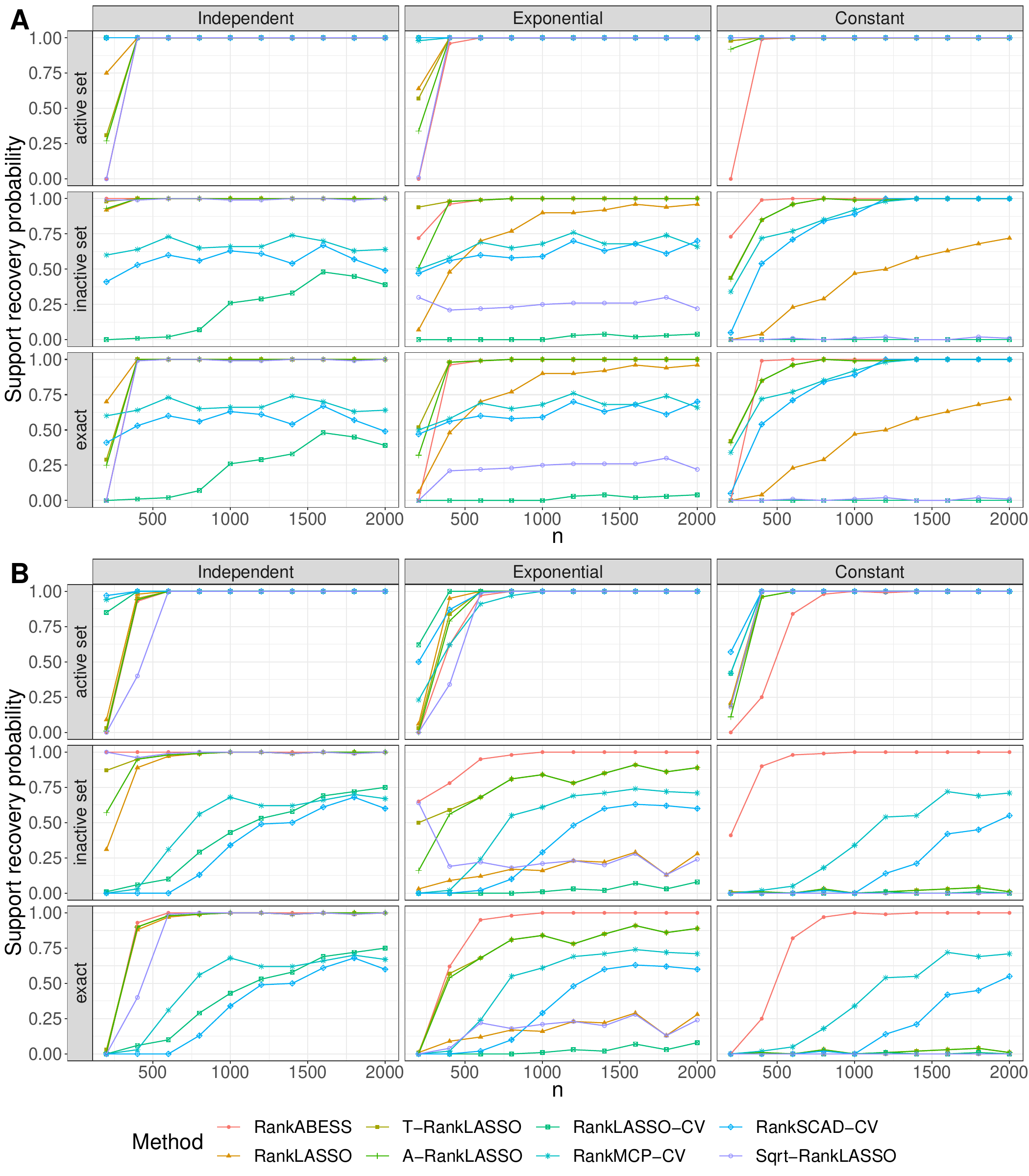}
\caption{{Performance of 8 methods on Model (a).} Performance for subset selection under different covariance structures and error distributions, measured by support recovery probabilities. Panel A for Gaussian distribution $ N(0,1) $, Panel B for Cauchy distribution $ t(1) $.}
\label{fig:modela}
\end{figure}

\begin{figure}
\centering
\includegraphics[width=0.9\linewidth]{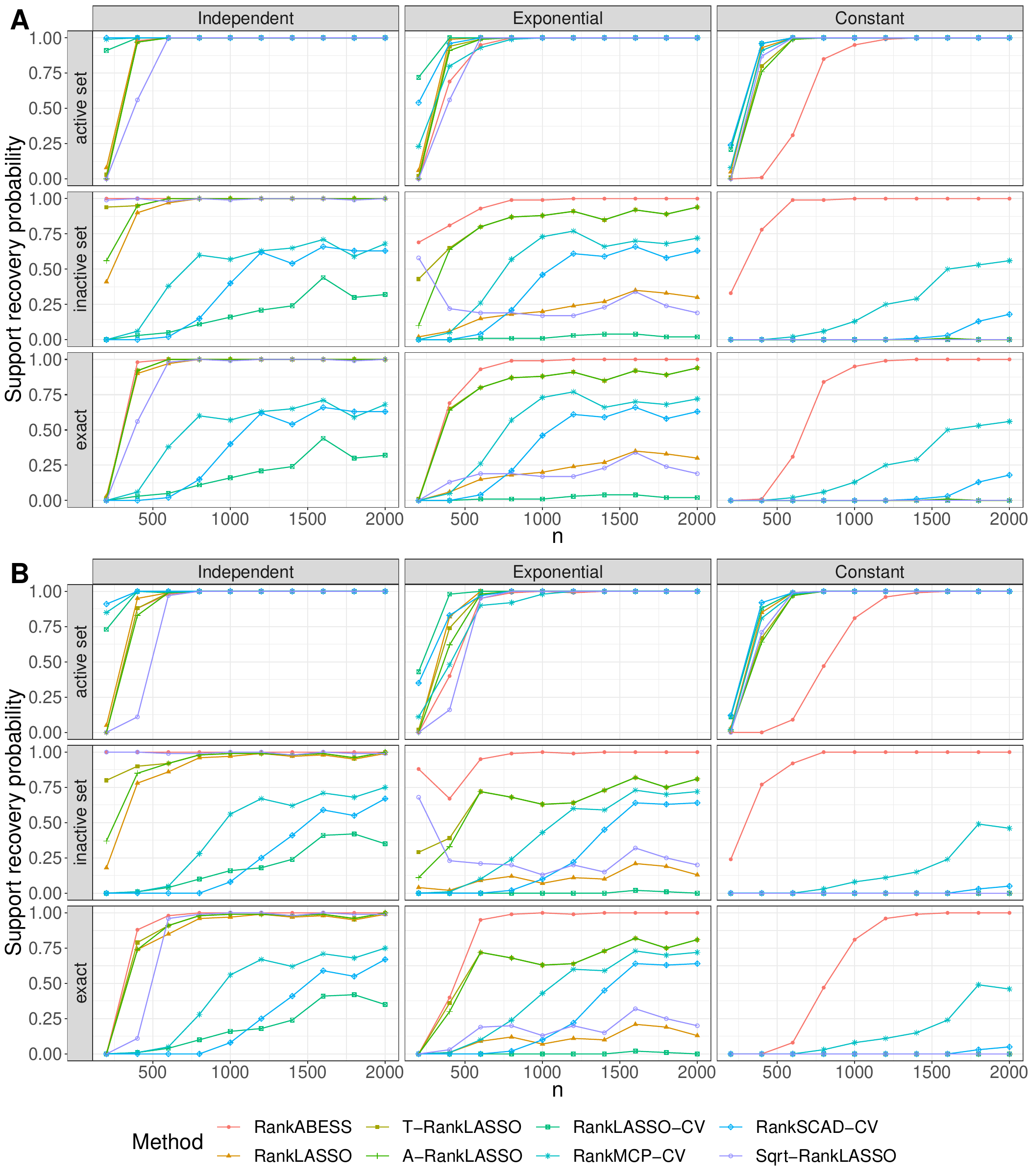}
\caption{{Performance of 8 methods on Model (b)}. Performance for subset selection under different covariance structures and error distributions, measured by support recovery probabilities. Panel A for Gaussian distribution $ N(0,1) $, Panel B for Cauchy distribution $ t(1)$.}
\label{fig:modelb}
\end{figure}

In this section, we empirically demonstrate the statistical and computational properties of our proposed method which is compared with some
state-of-the-art variable selection methods for SIMs. Specifically, the following methods are compared.
\begin{enumerate}[leftmargin=*]
\item \textbf{RankABESS}: implemented with R package \textit{abess} \citep{zhu2022abess} The GIC is applied for support size selection.
\item \textbf{RankMCP-CV \& RankSCAD-CV}: implemented with R package \textit{ncvreg}, A 10-fold cross-validation (CV) is applied for tuning parameter selection. See \cite{wang2012non}.
\item \textbf{RankLASSO-CV}: implemented with R package \textit{glmnet}, 10-fold CV is applied for tuning parameter selection. See \cite{wang2015distribution}.
\item \textbf{RankLASSO}: implemented with R package \textit{glmnet}. The parameter tuning follows
the method in \cite{rejchel2020rank}.
\item \textbf{T-RankLASSO \& A-RankLASSO}: thresholded and adaptive RankLASSO, implemented with R package \textit{glmnet}. The parameter tuning follows the method in \cite{rejchel2020rank}.
\item { \textbf{Sqrt-RankLASSO:} square-root RankLASSO. While this variant has not been previously studied, we implement it using the square-root LASSO code from Github\footnote{Sqrt-LASSO implementation, \url{https://github.com/jstriaukas/sqrt_lasso_rcpp}} and include it in our simulations, as it was mentioned by the reviewer. The tuning parameter is selected according to the recommendation of \cite{10.1093/biomet/asr043}.}
\end{enumerate}
For a fixed dimension $ p=2000 $, we increase the sample size from 200 to 2000 with equal step size and investigate the subset selection performance of different methods. The sparsity level is fixed as $ 10 $. We set the true regression coefficient $ \boldsymbol{\beta}^{\star} $ as follows. Ten elements equal $2$ and the
others are all $0$. The $10$ nonzero coefficients are equally spaced between the $10$-th and
$200$-th elements. 
The predictors are independently sampled from the multivariate Gaussian distribution $ \mathcal{M V N}(0, \Sigma) $, where $ \Sigma $ is a covariance matrix with three possible choices: the independent structure ($ \Sigma = I $), the exponential structure ($\Sigma_{i j}=\rho^{I(i \neq j)}$, where $ \rho = 0.8$), and the constant structure ($\Sigma_{i j}=\rho$,  where $ \rho = 0.2 $).
Then we independently draw the error from either the standard Gaussian distribution $ N(0,1) $ or the standard Cauchy distribution $ t(1) $ and generate response variables from the following two models:
\begin{enumerate}
\item[(a)]  $ y_i=\boldsymbol{x}_i^{\top} \boldsymbol{b} + \boldsymbol{e}_i $
\item[(b)] $ y_i = \exp (\boldsymbol{x}_i^{\top} \boldsymbol{b}) + \boldsymbol{e}_i$.
\end{enumerate}
It is noteworthy that RankABESS can also be applied for general single index models, where the error may not be additive with the linear predictor. Additional simulation results for such scenarios can be found in Section A of the appendix.
Performance for subset selection is measured by support recovery probability defined as follows,
\begin{enumerate}
\item[(i)] Recovery probability for the active set: $\mathbb{P}(\mathcal{A}^{\star} \subseteq \widehat{\mathcal{A}})$.
\item[(ii)] Recovery probability for the inactive set: $\mathbb{P}((\mathcal{A}^{\star})^c \subseteq \widehat{\mathcal{A}}^c)$.
\item[(iii)] Exact support recovery probability: $\mathbb{P}(\mathcal{A}^{\star}= \widehat{\mathcal{A}})$.
\end{enumerate}
All experiments are based on 100 synthetic datasets. Simulation results for Models (a) and (b) are presented in Figures \ref{fig:modela} and \ref{fig:modelb}, respectively. 

\begin{figure}[t]
\centering
\includegraphics[width=0.9 \linewidth]{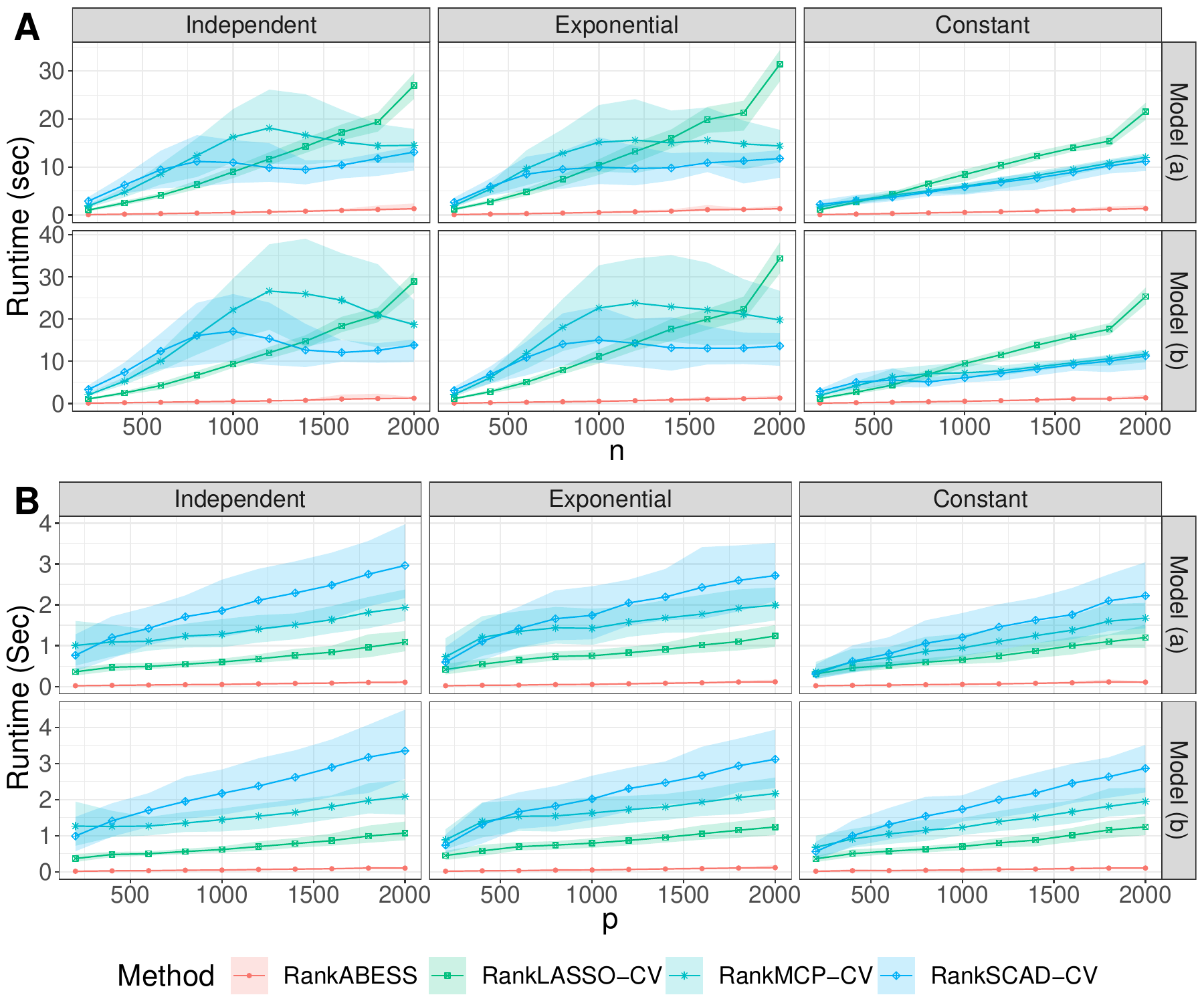}
\caption{Runtime (in seconds) of RankABESS and CV-based methods in different settings (A for runtime versus $ n $, B for runtime versus $ p $). The average runtime of 100 replications is recorded. The colored ribbons cover 5\%-95\% quantiles of runtime for each method.}
\label{fig:runtime}
\end{figure}

\noindent\textbf{Model (a).} Under model (a), as depicted by Figure \ref{fig:modela}A, RankABESS performs competitively in all cases. All methods can cover the active set quite well. However, for exact recovery, only RankABESS, A-RankLASSO and T-RankLASSO achieve consistency in all three $\Sigma$ structures. RankLASSO-CV performs the worst, while RankMCP-CV and RankSCAD-CV perform slightly better for very small $n$ but fail to achieve subset selection consistency as the sample size increases. Since CV procedures provide no guarantees for model selection consistency, these CV-based methods include irrelevant variables occasionally.  It is noteworthy that although RankLASSO have similar support recovery performance with A-RankLASSO and T-RankLASSO under the independent $\Sigma$ structure, it fails to achieve consistency under the correlated $\Sigma$ structure. The advantage of RankABESS becomes more apparent in Figure \ref{fig:modela}B, where the error distribution is heavy-tailed. In this scenario, A-RankLASSO and T-RankLASSO can only recover the true support with probability 1 in the independent correlation $\Sigma$ structure, while RankABESS achieves consistency in all settings. {Finally, we can see that Sqrt-RankLASSO is generally similar to RankLASSO on exact support recovery. Typically, compared with RankLASSO,  Sqrt-RankLASSO is less likely to select irrelevant variables but is generally surpassed in terms of identifying relevant variables.}

\noindent\textbf{Model (b).} Figure \ref{fig:modelb} presents the results under Model (b). The exponential link function challenges the performance of RankLASSO and its variants. However, RankABESS can still achieve model selection consistency for both Gaussian and Cauchy errors, although relatively larger sample sizes are required for high correlation settings.

In summary, our extensive simulation results demonstrate that RankABESS outperforms
other methods for the best-subset selection in high-dimensional SIMs, achieving the subset selection consistency across various settings, particularly in scenarios with high correlation, heavy-tailed error, and nonlinear link functions. 

\paragraph{Runtime.} Using the Cauchy error case as an illustration, we assess the scalability of RankABESS
versus the other methods in terms of the runtime. Three CV-based methods are chosen as the benchmark. Figure \ref{fig:runtime} displays the runtime: (A) a fixed dimension $p = 2000$ and sample sizes ranged from $ 200 $ to $ 2000 $; and (B) with a fixed sample size $n = 200$, the dimension increases  from $ 200 $ to $ 2000 $. RankABESS is overwhelmingly faster than other methods, which empirically demonstrates the computational efficiency of RankABESS. Moreover, the runtime of RankABESS grows in a linear pattern as the dimension increases when the sample size is fixed, and vice versa. This coincides with Theorem \ref{thm:time}, confirming that our algorithm achieves a polynomial complexity.

\begin{figure}[htbp]
\centering
\vspace*{-10pt}
\includegraphics[width=0.75\linewidth]{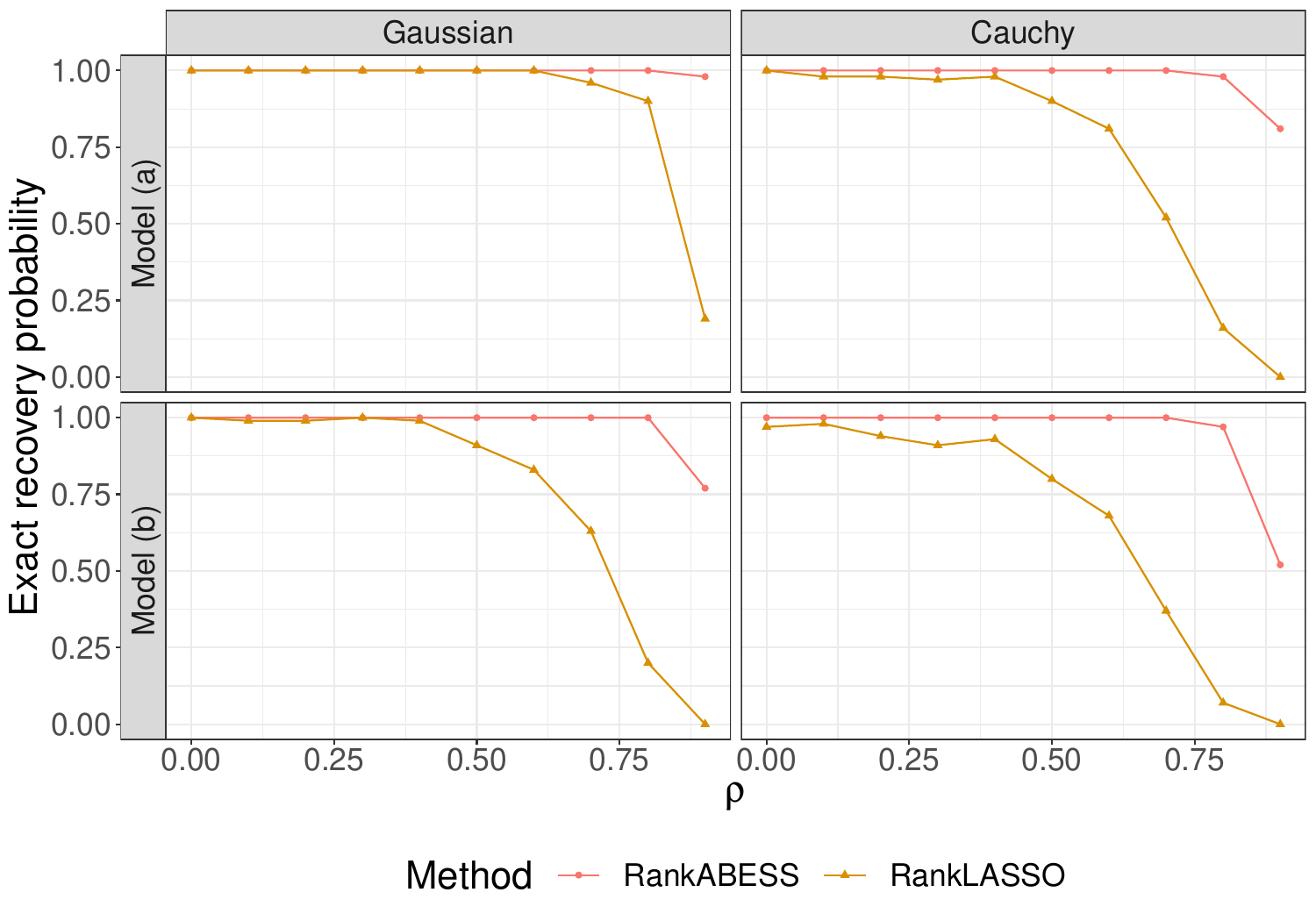}
\vspace*{-6pt}
\caption{{The exact support recovery probability when correlation strength $\rho$ increases. Each row corresponds to one single index model, and each column indicates one distribution of random noises.}}\label{fig:cor}
\end{figure}

{\paragraph{Effect of correlation strength.} In this part, we examine how the probability of exact support recovery changes as the correlation strength ($\rho$) increases. Without loss of generality, we focus on the exponential correlation scenario and compare RankABESS with RankLASSO to illustrate that RankABESS achieves accurate recovery under less restrictive assumptions regarding the design matrix. The experimental results are displayed in Figure~\ref{fig:cor}. As evident from the figure, RankABESS recovers the true support with probability one, even when the correlation level reaches $\rho = 0.7$, regardless of the link function form or the distribution of the random noise. In contrast, under the same correlation level ($\rho = 0.7$), the probability of RankLASSO accurately recovering the true support declines visibly, and may fall below 40\%.}

\section{Conclusion and Discussion} \label{sec:discuss}

In this paper, we propose a novel algorithm, the RankABESS, for the best-subset selection and index direction estimation in high-dimensional SIMs. To summarize, our contribution is several-fold. First, by adopting the simple rank-based transformation, our method is free of assumptions on the error distribution or the link function and hence has broader applications. Specifically, RankABESS can deal with not only a wide range of models, including generalized linear models, and classical single index models, but also heavy-tailed errors. Second, We rigorously prove that our algorithm can recover the true model for a given model size and that the true model size can be selected under the GIC with high probability, which is further demonstrated by empirical studies. Our proof of the algorithmic properties differs from that in \citet{zhu2020polynomial}, owing to new theoretical challenges posed by the dependence among the ranking responses, randomness of predictors, and the absence of assumptions. Third, our algorithm is computationally efficient and capable of high-dimensional datasets in practice. Notably, the scalability of our algorithm is theoretically justified as we rigorously prove that the algorithm has polynomial complexity. { To the best of our knowledge, this is the first polynomial algorithm that directly tackles the best-subset selection in high-dimensional SIMs.}

Our method has notable advantages over several existing methods. Firstly, it enjoys desirable theoretical properties, similar to the splicing algorithm for linear models or generalized linear models \citep{zhu2020polynomial}, with broader applications. Another recent development in general sparsity constraint optimization is greedy support pursuit (GRASP, \citet{bahmani2011greedy}). They require a {Stable Restricted Hessian or Stable Restricted Linearization condition on the loss function, which imposes constraints on $\boldsymbol{b}$} in addition to those on $X$ and may be stringent in some cases. A weaker but similar condition is also imposed in \cite{zhu2023best}. In contrast, such constraints on $\boldsymbol{b}$ are not required for RankABESS. Though both methods can be applied to generalized linear models, RankABESS is simpler in implementation, while GRASP requires a $\ell_2$-regularization to avoid singularity in logistic regression, for example. Finally, RankABESS does not require any tricky tuning and has the oracle property with high probability, while for (modified) RankLASSO, open problems related to tuning parameter selection still exist, and biased estimators are obtained. 

A meaningful topic for future study is to extend our method for the best-subset of groups selection problem, which is useful to analyze variables with certain group structures and has drawn the interest of many researchers \citep{yuan2006model}. 
Recently, a splicing type algorithm has been developed for this problem in linear models \citep{zhang2022splicing}. This paves the way to study the best-subset of groups selection in SIMs.
Another future direction to explore is to alleviate assumptions on the predictors. Although achieving robustness to the error distribution, we pay the cost of imposing the linearity and subgaussianity conditions on the predictors.
On the other hand, robust variable selection procedures without restrictive conditions on the predictors have been developed in \cite{wang2013robust} and \cite{wang2020tuning} for linear models,
which are inspiring for further studies. 
{In more general settings of empirical risk minimization, \cite{hsu2016loss} provide an alternative generic proposal for heavy-tailed data.}
	
	\bibliographystyle{Chicago}
	\bibliography{SIM_arxiv}

\begin{thebibliography}{}

\bibitem[\protect\citeauthoryear{Alquier and Biau}{Alquier and
  Biau}{2013}]{alquier2013sparse}
Alquier, P. and G.~Biau (2013).
\newblock Sparse single-index model.
\newblock {\em Journal of Machine Learning Research\/}~{\em 14\/}(1).

\bibitem[\protect\citeauthoryear{Bahmani, Raj, and Boufounos}{Bahmani
  et~al.}{2013}]{bahmani2011greedy}
Bahmani, S., B.~Raj, and P.~T. Boufounos (2013).
\newblock Greedy sparsity-constrained optimization.
\newblock {\em Journal of Machine Learning Research\/}~{\em 14\/}(25),
  807--841.

\bibitem[\protect\citeauthoryear{Belloni and Chernozhukov}{Belloni and
  Chernozhukov}{2013}]{10.3150/11-BEJ410}
Belloni, A. and V.~Chernozhukov (2013).
\newblock {Least squares after model selection in high-dimensional sparse
  models}.
\newblock {\em Bernoulli\/}~{\em 19\/}(2), 521 -- 547.

\bibitem[\protect\citeauthoryear{Belloni, Chernozhukov, and Wang}{Belloni
  et~al.}{2011}]{10.1093/biomet/asr043}
Belloni, A., V.~Chernozhukov, and L.~Wang (2011, 12).
\newblock Square-root lasso: pivotal recovery of sparse signals via conic
  programming.
\newblock {\em Biometrika\/}~{\em 98\/}(4), 791--806.

\bibitem[\protect\citeauthoryear{Bertsimas, Pauphilet, and Van~Parys}{Bertsimas
  et~al.}{2020}]{Bertsimas_2020}
Bertsimas, D., J.~Pauphilet, and B.~Van~Parys (2020, November).
\newblock Sparse regression: Scalable algorithms and empirical performance.
\newblock {\em Statistical Science\/}~{\em 35\/}(4).

\bibitem[\protect\citeauthoryear{Brillinger}{Brillinger}{2012}]{Brillinger2012}
Brillinger, D.~R. (2012).
\newblock {\em A Generalized Linear Model With ``Gaussian'' Regressor
  Variables}, pp.\  589--606.
\newblock Springer New York.

\bibitem[\protect\citeauthoryear{Cheng, Zeng, and Zhu}{Cheng
  et~al.}{2017}]{10.1214/17-EJS1329}
Cheng, L., P.~Zeng, and Y.~Zhu (2017).
\newblock {BS-SIM: An effective variable selection method for high-dimensional
  single index model}.
\newblock {\em Electronic Journal of Statistics\/}~{\em 11\/}(2), 3522 -- 3548.

\bibitem[\protect\citeauthoryear{Eftekhari, Banerjee, and Ritov}{Eftekhari
  et~al.}{2021}]{eftekhari2021inference}
Eftekhari, H., M.~Banerjee, and Y.~Ritov (2021).
\newblock Inference in high-dimensional single-index models under symmetric
  designs.
\newblock {\em Journal of Machine Learning Research\/}~{\em 22\/}(27), 1--63.

\bibitem[\protect\citeauthoryear{Fan and Li}{Fan and
  Li}{2001}]{fan2001variable}
Fan, J. and R.~Li (2001).
\newblock Variable selection via nonconcave penalized likelihood and its oracle
  properties.
\newblock {\em Journal of the American Statistical Association\/}~{\em
  96\/}(456), 1348--1360.

\bibitem[\protect\citeauthoryear{Fan, Yang, and Yu}{Fan
  et~al.}{2022}]{fan2022understanding}
Fan, J., Z.~Yang, and M.~Yu (2022).
\newblock Understanding implicit regularization in over-parameterized single
  index model.
\newblock {\em Journal of the American Statistical Association\/}~{\em 0\/}(0),
  1--14.

\bibitem[\protect\citeauthoryear{Fan and Tang}{Fan and
  Tang}{2013}]{fan2013tuning}
Fan, Y. and C.~Y. Tang (2013).
\newblock Tuning parameter selection in high dimensional penalized likelihood.
\newblock {\em Journal of the Royal Statistical Society. Series B (Statistical
  Methodology)\/}~{\em 75\/}(3), 531--552.

\bibitem[\protect\citeauthoryear{Ganti, Rao, Balzano, Willett, and Nowak}{Ganti
  et~al.}{2017}]{Ganti_Rao_Balzano_Willett_Nowak_2017}
Ganti, R., N.~Rao, L.~Balzano, R.~Willett, and R.~Nowak (2017).
\newblock On learning high dimensional structured single index models.
\newblock {\em Proceedings of the AAAI Conference on Artificial
  Intelligence\/}~{\em 31\/}(1).

\bibitem[\protect\citeauthoryear{Hall and Li}{Hall and
  Li}{1993}]{hall1993almost}
Hall, P. and K.-C. Li (1993).
\newblock {On almost Linearity of Low Dimensional Projections from High
  Dimensional Data}.
\newblock {\em The Annals of Statistics\/}~{\em 21\/}(2), 867 -- 889.

\bibitem[\protect\citeauthoryear{Horowitz and Härdle}{Horowitz and
  Härdle}{1996}]{horowitz1996direct}
Horowitz, J.~L. and W.~Härdle (1996).
\newblock Direct semiparametric estimation of single-index models with discrete
  covariates.
\newblock {\em Journal of the American Statistical Association\/}~{\em
  91\/}(436), 1632--1640.

\bibitem[\protect\citeauthoryear{Hsu and Sabato}{Hsu and
  Sabato}{2016}]{hsu2016loss}
Hsu, D. and S.~Sabato (2016).
\newblock Loss minimization and parameter estimation with heavy tails.
\newblock {\em Journal of Machine Learning Research\/}~{\em 17\/}(18), 1--40.

\bibitem[\protect\citeauthoryear{Huang, Jiao, Liu, and Lu}{Huang
  et~al.}{2018}]{huang2018constructive}
Huang, J., Y.~Jiao, Y.~Liu, and X.~Lu (2018).
\newblock A constructive approach to $l_0$ penalized regression.
\newblock {\em Journal of Machine Learning Research\/}~{\em 19\/}(10), 1--37.

\bibitem[\protect\citeauthoryear{Ichimura}{Ichimura}{1993}]{ichimura1993semiparametric}
Ichimura, H. (1993).
\newblock Semiparametric least squares (sls) and weighted sls estimation of
  single-index models.
\newblock {\em Journal of Econometrics\/}~{\em 58}, 71--120.

\bibitem[\protect\citeauthoryear{Kakade, Kanade, Shamir, and Kalai}{Kakade
  et~al.}{2011}]{NIPS2011_30bb3825}
Kakade, S.~M., V.~Kanade, O.~Shamir, and A.~Kalai (2011).
\newblock Efficient learning of generalized linear and single index models with
  isotonic regression.
\newblock In J.~Shawe-Taylor, R.~Zemel, P.~Bartlett, F.~Pereira, and
  K.~Weinberger (Eds.), {\em Advances in Neural Information Processing
  Systems}, Volume~24. Curran Associates, Inc.

\bibitem[\protect\citeauthoryear{Kong and Xia}{Kong and
  Xia}{2007}]{kong2007variable}
Kong, E. and Y.~Xia (2007).
\newblock Variable selection for the single-index model.
\newblock {\em Biometrika\/}~{\em 94\/}(1), 217--229.

\bibitem[\protect\citeauthoryear{Li}{Li}{1991}]{li1991sliced}
Li, K.-C. (1991).
\newblock Sliced inverse regression for dimension reduction.
\newblock {\em Journal of the American Statistical Association\/}~{\em
  86\/}(414), 316--327.

\bibitem[\protect\citeauthoryear{Li and Duan}{Li and
  Duan}{1989}]{li1989regression}
Li, K.-C. and N.~Duan (1989).
\newblock {Regression analysis under link violation}.
\newblock {\em The Annals of Statistics\/}~{\em 17\/}(3), 1009 -- 1052.

\bibitem[\protect\citeauthoryear{Lounici}{Lounici}{2008}]{10.1214/08-EJS177}
Lounici, K. (2008).
\newblock {Sup-norm convergence rate and sign concentration property of Lasso
  and Dantzig estimators}.
\newblock {\em Electronic Journal of Statistics\/}~{\em 2\/}(none), 90 -- 102.

\bibitem[\protect\citeauthoryear{Luo and Ghosal}{Luo and
  Ghosal}{2016}]{luo2016forward}
Luo, S. and S.~Ghosal (2016).
\newblock Forward selection and estimation in high dimensional single index
  models.
\newblock {\em Statistical Methodology\/}~{\em 33}, 172--179.

\bibitem[\protect\citeauthoryear{Meinshausen and B{\"u}hlmann}{Meinshausen and
  B{\"u}hlmann}{2006}]{10.1214/009053606000000281}
Meinshausen, N. and P.~B{\"u}hlmann (2006).
\newblock {High-dimensional graphs and variable selection with the Lasso}.
\newblock {\em The Annals of Statistics\/}~{\em 34\/}(3), 1436 -- 1462.

\bibitem[\protect\citeauthoryear{Na, Yang, Wang, and Kolar}{Na
  et~al.}{2019}]{JMLR:v20:18-705}
Na, S., Z.~Yang, Z.~Wang, and M.~Kolar (2019).
\newblock High-dimensional varying index coefficient models via stein's
  identity.
\newblock {\em Journal of Machine Learning Research\/}~{\em 20\/}(152), 1--44.

\bibitem[\protect\citeauthoryear{Natarajan}{Natarajan}{1995}]{natarajan1995sparse}
Natarajan, B.~K. (1995).
\newblock Sparse approximate solutions to linear systems.
\newblock {\em SIAM Journal on Computing\/}~{\em 24\/}(2), 227--234.

\bibitem[\protect\citeauthoryear{Neykov, Liu, and Cai}{Neykov
  et~al.}{2016}]{neykov2016l1}
Neykov, M., J.~S. Liu, and T.~Cai (2016).
\newblock L1-regularized least squares for support recovery of high dimensional
  single index models with gaussian designs.
\newblock {\em Journal of Machine Learning Research\/}~{\em 17\/}(87), 1--37.

\bibitem[\protect\citeauthoryear{Plan and Vershynin}{Plan and
  Vershynin}{2016}]{7378952}
Plan, Y. and R.~Vershynin (2016).
\newblock The generalized lasso with non-linear observations.
\newblock {\em IEEE Transactions on Information Theory\/}~{\em 62\/}(3),
  1528--1537.

\bibitem[\protect\citeauthoryear{Radchenko}{Radchenko}{2015}]{radchenko2015high}
Radchenko, P. (2015).
\newblock High dimensional single index models.
\newblock {\em Journal of Multivariate Analysis\/}~{\em 139}, 266--282.

\bibitem[\protect\citeauthoryear{Ravikumar, Wainwright, Raskutti, and
  Yu}{Ravikumar et~al.}{2011}]{ravikumar2011high}
Ravikumar, P., M.~J. Wainwright, G.~Raskutti, and B.~Yu (2011).
\newblock {High-dimensional covariance estimation by minimizing
  $\ell_1$-penalized log-determinant divergence}.
\newblock {\em Electronic Journal of Statistics\/}~{\em 5}, 935 -- 980.

\bibitem[\protect\citeauthoryear{Rejchel and Bogdan}{Rejchel and
  Bogdan}{2020}]{rejchel2020rank}
Rejchel, W. and M.~Bogdan (2020).
\newblock {Rank-based Lasso - efficient methods for high-dimensional robust
  model selection}.
\newblock {\em Journal of Machine Learning Research\/}~{\em 21\/}(244), 1--47.

\bibitem[\protect\citeauthoryear{Sheng and Yin}{Sheng and
  Yin}{2013}]{sheng2013direction}
Sheng, W. and X.~Yin (2013).
\newblock Direction estimation in single-index models via distance covariance.
\newblock {\em Journal of Multivariate Analysis\/}~{\em 122}, 148--161.

\bibitem[\protect\citeauthoryear{Stein, Diaconis, Holmes, and Reinert}{Stein
  et~al.}{2004}]{stein2004use}
Stein, C., P.~Diaconis, S.~Holmes, and G.~Reinert (2004).
\newblock Use of exchangeable pairs in the analysis of simulations.
\newblock {\em Lecture Notes-Monograph Series\/}~{\em 46}, 1--25.

\bibitem[\protect\citeauthoryear{Tibshirani}{Tibshirani}{1996}]{tibshirani1996regression}
Tibshirani, R. (1996).
\newblock Regression shrinkage and selection via the lasso.
\newblock {\em Journal of the Royal Statistical Society. Series B
  (Methodological)\/}~{\em 58\/}(1), 267--288.

\bibitem[\protect\citeauthoryear{Wang, Peng, Bradic, Li, and Wu}{Wang
  et~al.}{2020}]{wang2020tuning}
Wang, L., B.~Peng, J.~Bradic, R.~Li, and Y.~Wu (2020).
\newblock A tuning-free robust and efficient approach to high-dimensional
  regression.
\newblock {\em Journal of the American Statistical Association\/}~{\em
  115\/}(532), 1700--1714.

\bibitem[\protect\citeauthoryear{Wang, Xu, and Zhu}{Wang
  et~al.}{2012}]{wang2012non}
Wang, T., P.-R. Xu, and L.-X. Zhu (2012).
\newblock Non-convex penalized estimation in high-dimensional models with
  single-index structure.
\newblock {\em Journal of Multivariate Analysis\/}~{\em 109}, 221--235.

\bibitem[\protect\citeauthoryear{Wang and Zhu}{Wang and
  Zhu}{2015}]{wang2015distribution}
Wang, T. and L.~Zhu (2015).
\newblock A distribution-based lasso for a general single-index model.
\newblock {\em Science China Mathematics\/}~{\em 58}, 109--130.

\bibitem[\protect\citeauthoryear{Wang, Jiang, Huang, and Zhang}{Wang
  et~al.}{2013}]{wang2013robust}
Wang, X., Y.~Jiang, M.~Huang, and H.~Zhang (2013).
\newblock Robust variable selection with exponential squared loss.
\newblock {\em Journal of the American Statistical Association\/}~{\em
  108\/}(502), 632--643.

\bibitem[\protect\citeauthoryear{Wei, Yang, and Wang}{Wei
  et~al.}{2019}]{pmlr-v97-wei19b}
Wei, X., Z.~Yang, and Z.~Wang (2019).
\newblock On the statistical rate of nonlinear recovery in generative models
  with heavy-tailed data.
\newblock In {\em Proceedings of the 36th International Conference on Machine
  Learning}, Volume~97, pp.\  6697--6706.

\bibitem[\protect\citeauthoryear{Xia, Tong, Li, and Zhu}{Xia
  et~al.}{2002}]{xia2009adaptive}
Xia, Y., H.~Tong, W.~K. Li, and L.-X. Zhu (2002).
\newblock An adaptive estimation of dimension reduction space.
\newblock {\em Journal of the Royal Statistical Society: Series B (Statistical
  Methodology)\/}~{\em 64\/}(3), 363--410.

\bibitem[\protect\citeauthoryear{Yang, Balasubramanian, and Liu}{Yang
  et~al.}{2017}]{yang2017high}
Yang, Z., K.~Balasubramanian, and H.~Liu (2017).
\newblock High-dimensional non-{G}aussian single index models via thresholded
  score function estimation.
\newblock In {\em Proceedings of the 34th International Conference on Machine
  Learning}, Volume~70, pp.\  3851--3860.

\bibitem[\protect\citeauthoryear{Yuan and Lin}{Yuan and
  Lin}{2006}]{yuan2006model}
Yuan, M. and Y.~Lin (2006).
\newblock Model selection and estimation in regression with grouped variables.
\newblock {\em Journal of the Royal Statistical Society: Series B (Statistical
  Methodology)\/}~{\em 68\/}(1), 49--67.

\bibitem[\protect\citeauthoryear{Zhang}{Zhang}{2010}]{10.1214/09-AOS729}
Zhang, C.-H. (2010).
\newblock {Nearly unbiased variable selection under minimax concave penalty}.
\newblock {\em The Annals of Statistics\/}~{\em 38\/}(2), 894 -- 942.

\bibitem[\protect\citeauthoryear{Zhang and Huang}{Zhang and
  Huang}{2008}]{zhang2008sparsity}
Zhang, C.-H. and J.~Huang (2008).
\newblock {The sparsity and bias of the Lasso selection in high-dimensional
  linear regression}.
\newblock {\em The Annals of Statistics\/}~{\em 36\/}(4), 1567 -- 1594.

\bibitem[\protect\citeauthoryear{Zhang and Yin}{Zhang and
  Yin}{2015}]{zhang2015direction}
Zhang, N. and X.~Yin (2015).
\newblock Direction estimation in single-index regressions via hilbert-schmidt
  independence criterion.
\newblock {\em Statistica Sinica\/}~{\em 25}, 743--758.

\bibitem[\protect\citeauthoryear{Zhang, Zhu, Zhu, and Wang}{Zhang
  et~al.}{2023}]{zhang2022splicing}
Zhang, Y., J.~Zhu, J.~Zhu, and X.~Wang (2023).
\newblock A splicing approach to best subset of groups selection.
\newblock {\em INFORMS Journal on Computing\/}~{\em 35\/}(1), 104--119.

\bibitem[\protect\citeauthoryear{Zhong, Liu, and Ma}{Zhong
  et~al.}{2017}]{zhong2018variable}
Zhong, W., X.~Liu, and S.~Ma (2017).
\newblock Variable selection and direction estimation for single-index models
  via {DC-TGDR} method.
\newblock {\em Statistics and Its Interface\/}~{\em 11\/}(1), 169--181.

\bibitem[\protect\citeauthoryear{Zhu, Wang, Hu, Huang, Jiang, Zhang, Lin, and
  Zhu}{Zhu et~al.}{2022}]{zhu2022abess}
Zhu, J., X.~Wang, L.~Hu, J.~Huang, K.~Jiang, Y.~Zhang, S.~Lin, and J.~Zhu
  (2022).
\newblock abess: A fast best-subset selection library in {Python} and {R}.
\newblock {\em Journal of Machine Learning Research\/}~{\em 23\/}(202), 1--7.

\bibitem[\protect\citeauthoryear{Zhu, Wen, Zhu, Zhang, and Wang}{Zhu
  et~al.}{2020}]{zhu2020polynomial}
Zhu, J., C.~Wen, J.~Zhu, H.~Zhang, and X.~Wang (2020).
\newblock A polynomial algorithm for best-subset selection problem.
\newblock {\em Proceedings of the National Academy of Sciences\/}~{\em
  117\/}(52), 33117--33123.

\bibitem[\protect\citeauthoryear{Zhu, Zhu, Tang, Chen, Lin, and Wang}{Zhu
  et~al.}{2023}]{zhu2023best}
Zhu, J., J.~Zhu, B.~Tang, X.~Chen, H.~Lin, and X.~Wang (2023).
\newblock Best-subset selection in generalized linear models: A fast and
  consistent algorithm via splicing technique.
\newblock {\em arXiv preprint\/}.

\bibitem[\protect\citeauthoryear{Zou}{Zou}{2006}]{zou2006adaptive}
Zou, H. (2006).
\newblock The adaptive lasso and its oracle properties.
\newblock {\em Journal of the American Statistical Association\/}~{\em
  101\/}(476), 1418--1429.

\end{thebibliography}
\end{document}